\pdfoutput=1
\documentclass[11pt]{article}
\usepackage{authblk}

\usepackage[]{acl}

\usepackage{times}
\usepackage{latexsym}
\usepackage{graphicx}
\usepackage{caption}
\usepackage{booktabs}
\usepackage{amsmath}
\usepackage{amsfonts}
\usepackage{xurl}

\usepackage[T1]{fontenc}
\usepackage[utf8]{inputenc}

\usepackage{cleveref}
\crefname{section}{\S}{\S\S}

\usepackage{microtype}

\title{Collaboration or Corporate Capture? Quantifying NLP's Reliance on Industry Artifacts and Contributions\vspace{3pt}}

\author[1,2*]{Will Aitken}
\author[3]{\hspace{1pt}Mohamed Abdalla\hspace{1pt}}
\author[1,2]{\hspace{1pt}Karen Rudie}
\author[4,5]{\hspace{1pt}Catherine Stinson}
\affil[1]{Department of Electrical and Computer Engineering, Queen’s University\vspace{-3pt}}
\affil[2]{Ingenuity Labs Research Institute, Queen's University\vspace{-3pt}}
\affil[3]{University of Alberta\vspace{-3pt}}
\affil[4]{School of Computing, Queen's University\vspace{-3pt}}
\affil[5]{Department of Philosophy, Queen's University\vspace{-3pt}}

\affil[ ]{\texttt{\{will.aitken, karen.rudie, c.stinson\}@queensu.ca, mabdall2@ualberta.ca}}

\begin{document}
\maketitle
\begingroup\def\thefootnote{*}\footnotetext{Corresponding Author.}\endgroup
% \begingroup\def\thefootnote{*}\footnotetext{Corresponding author.}\endgroup
% \renewcommand{\thefootnote}{\arabic{footnote}}
\begin{abstract} 
%% Mohamed Version of things. I thought the addition of a sentence at the end in the original version was too abrupt so I tried to massage it into the text.
Impressive performance of pre-trained models has garnered public attention and made news headlines in recent years. Almost always, these models are produced by or in collaboration with industry. Using them is critical for competing on natural language processing (NLP) benchmarks and correspondingly to stay relevant in NLP research. We surveyed 100 papers published at EMNLP 2022 to determine the degree to which researchers rely on industry models, other artifacts, and contributions to publish in prestigious NLP venues and found that the ratio of their citation is at least three times greater than what would be expected.  Our work serves as a scaffold to enable future researchers to more accurately address whether: 1) Collaboration with industry is still collaboration in the absence of an alternative or 2) if NLP inquiry has been captured by the motivations and research direction of private corporations. 
\end{abstract}

\section{Introduction}

Natural Language Processing (NLP) has seen rapid growth in recent years, attracting interest from large technology companies. Expediting the exploitation of emergent technologies is critical to industry success and this strategy has been applied to NLP, as is evident with the current influx of large language models (LLM), chatbots, and evaluation tools. These newfound business opportunities have brought industry to the forefront of NLP research. Industry presence in the Association of Computational Linguistics (ACL) Anthology has correspondingly increased 180\% from 2017 to 2022 \citep{abdalla-etal-2023-elephant}. The research community has taken notice: "industry races ahead of academia" was the top takeaway of the 2023 Artificial Intelligence (AI) Index Report \citep{index-report} and 82\% of NLP community survey respondents agreed with the statement "Industry will produce the most widely-cited research" \citep{michael-etal-2023-nlp}. Public investment in AI has failed to keep up with industry: in 2021, Google's funding of their subsidiary, DeepMind, alone surpassed the entirety of US non-military AI funding \citep{science-industry-numbers}. %Industry has invested substantial resources in AI and their presence and influence is likely to continue to rise.

\begin{figure}
    \centering
    \resizebox{\linewidth}{!}{
        \includegraphics{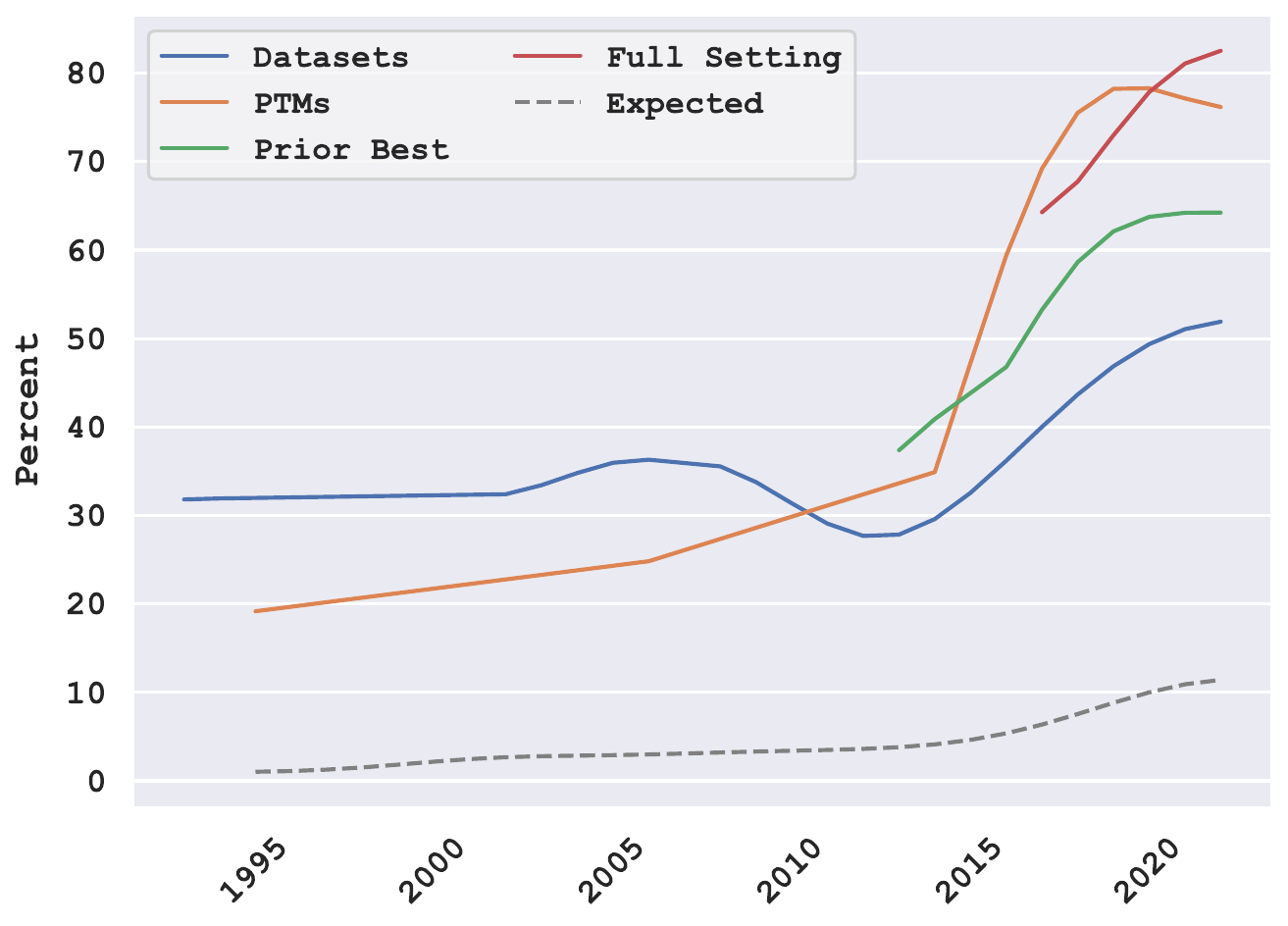}
    }
    \caption{Percentage of surveyed EMNLP 2022 paper citations with industry affiliation smoothed with a 1D Gaussian Filter ($\sigma = 2.5$). The "Expected" line is the percentage of industry affiliation across the entire ACL anthology over time. One could expect cited papers of a given year to have the same degree of industry affiliation. The citations are split into the following types: Datasets, Pre-Trained Models, Prior Bests, and Full Setting scores (See \cref{sec:citation_def} for definitions of citation types).} 
    \label{fig:years}
\end{figure}

In parallel to increased industry presence, concerns have surfaced regarding the efficacy of research grounded on general purpose benchmark evaluation. \citet{grover-benchmarks} note that widely-cited Natural Language Understanding (NLU) datasets used to verify general understanding of language models are subject to scope and subjectivity limitations. Despite these documented flaws, performance on held-out benchmark test sets is still the \textit{de facto} standard used to establish publication validity (see \cref{sec:paper_type}). The consensus from NLP researchers interviewed in a recent qualitative study \citep{gururaja-etal-2023-build} was that ``improvements on benchmarks in NLP are the only results that are self-justifying to reviewers''. Furthermore, benchmark test sets favour institutes with higher computational budgets, since higher scores can sometimes be achieved via sheer computational power \citep{dodge-etal-2019-show}. 

%If the NLP community requires state-of-the-art (SOTA) benchmark performance to publish in top venues and if competing on such benchmarks favours those with more computing power

If substantial computer power is required to achieve state-of-the-art (SOTA) benchmark performance and thus publish in top venues \citep{dodge-etal-2019-show, reject-not-sota}, researchers with smaller budgets may find little room for contribution without relying on industry artifacts and contributions. 

In this work, we define reliance as needing something to a point such that, without it, one could not survive or be successful. Reliance enforces necessity which in turn minimizes the possibility of choosing an alternative. Instances of reliance have already been identified in NLP. For example, an NLP research community member notes regarding software tools: ``You’re not gonna just build your own system that’s gonna compete on these major benchmarks yourself. You have to start [with] the infrastructure that is there'' \cite{gururaja-etal-2023-build}.

Increased reliance on industry artifacts increases the risk of corporate capture -- a phenomenon where private industry uses its influence to affect research direction and findings (i.e., Science Capture), policy (i.e, Policy Capture), or media coverage (i.e., Media Capture) \cite{miller2010corporate}. Without provable intent (often obtained through litigation), or a counterfactual, it is exceedingly difficult to prove that `increased enmeshment of industry and academic research' \cite{facct-conflict} \textit{is} industry capture instead of mutually beneficial collaboration.

The purpose of this work is to quantify academia's increased reliance on industry artifacts. Since heavy reliance often co-occurs with capture, this is something we believe the field should study further. \textbf{We note that our work is not proof of industry capture.}
%To quantify whether this hypothetical relation is supported by data, 

In this study, we surveyed and collected metadata on 100 EMNLP 2022 papers. We found that across types of cited artifacts and contributions, the percentage that are industry affiliated is substantially higher than what would be expected relative to the collective ACL Anthology (Figure~\ref{fig:years}, \cref{sec:reliance}). Furthermore, the trend of industry affiliation has increased in recent years, culminating in the majority of all types of citations we recorded from 2022 being industry affiliated despite only 13.3\% of papers published that year having industry affiliation. 

%Our goal with this paper is to inform and disseminate resources to facilitate discourse regarding industry presence in NLP, and to discuss how viewpoints expressed in the existing literature might frame our findings with respect to the health of the AI research ecosystem. 

Along with determining the degree of reliance on industry (\cref{sec:reliance}) we designed our EMNLP 2022 survey to address the following additional questions:

\begin{enumerate}
    \item What proportion of papers have industry affiliation? (\cref{sec:affil_def})
    \item Are SOTA-pushing papers dominant? What other types of papers are being published and are their distributions different between academia and industry? (\cref{sec:paper_type})
    \item For new SOTA papers, by how much are industry and academia improving over the prior best? (\cref{sec:degree})
    \item What attributes of individual papers make them more or less favourable to achieving higher SOTA improvements? (\cref{sec:qual})
\end{enumerate}

The remainder of this paper is divided into related works (\cref{sec:related_works}), survey methods (\cref{sec:survey}), results from analyzing the survey data (\cref{sec:analysis}), and a corresponding discussion (\cref{sec:discussion}). The survey and analysis code are available (for research purposes only) at \url{https://github.com/Will-C-Aitken/collaboration-or-corporate-capture}.

\section{Related Works} \label{sec:related_works}

\subsection{Industry \& Scientific Research}
Undoubtedly, there are benefits to the presence of and increased collaboration with industry sponsors. \citet{gulbranson-ind-funding-and-performance} found that industry-funded academics have higher publication productivity and that their funding sources enable them to examine both more novel and more interesting research topics. The outputs of industry-funded science are often more application focused and the well-defined design rules associated with this line of research need not be considered any less epistemically virtuous than a more theoretical approach \citep{wilholt-industrial-merit}. In general, \citet{reconciling-anti-pro-industry} remarks that policy research tends to favour academia-industry ties while philosophy of science literature is more likely to condemn them.

The ideal of value-free science is largely considered unrealistic in the philosophy of science community. Some degree of bias is inevitable \citep{wilholt-preference-bias}.  %Nevertheless, some works propound that certain types of bias resulting from industry-influence are distinguishable and epistemically deficient, namely preference bias constrained within the context of research community conventions
However, some biases are more damaging to research integrity than others. Examples include preference bias---``when a research result unduly reflects the researchers’ preference for it over other possible results'' \citep{wilholt-preference-bias}--- and sponsorship bias that leads to using insufficient evidence to support dubious scientific claims \citep{evidential-account}.

The proliferation of industry presence in AI research has also garnered internal criticism. Concerns regarding academic independence and integrity have been voiced by \citet{abdalla-grey-hoodie} and \citet{whittaker-cold-war} by drawing parallels to tactics used by "Big Tobacco" and the US Military respectively to support their financial and political interests at the expense of public health and scientific transparency and soundness. 

Supporting AI Ethics research has been described as a tactic used by tech firms to gain social capital and foster a favourable public image \citep{phan-ethics-capital, steele-machinewash}. While this research can be beneficial, this practice can also have negative consequences. Some examples are how industry funding of digital rights civil society organizations can result in misrepresentation of public interests \citep{digital-rights} and how corporate capture of the ACM Fairness, Accountability, and Transparency (FAccT) conference led to conflicts of interest in the peer-review process \citep{facct-conflict}. 

\subsection{Benchmark Critiques}

The acceptance of a paper titled "What Will it Take to Fix Benchmarking in Natural Language Understanding" (\citealp{bowman-dahl-2021-will}) to NAACL 2020 is strongly indicative of widespread benchmark dissent. The authors provide criteria that they believe future benchmarks should strive for. \citet{grover-benchmarks} critique the generalizability of benchmarks by calling into question the validity of their constructs. Adjacently, through a microeconomics lens, \citet{ethayarajh-jurafsky-2020-utility} argue that benchmark leaderboards fail to adequately capture model utility. Despite its shortcomings, benchmarking remains pervasive because it makes model comparison, evaluation of results, and quantification of progress straightforward. 

\subsection{Computing Power}

The resurgence of NLP in recent years can in part be attributed to increased computing power. \citet{cost-of-nlp} estimate the cost of training an 11B-parameter model (a modest size by today's standards) to be well above \$1.3 Million USD, a cost likely prohibitive in most academic settings. As a result, from 2010 to 2021, industry's share of large AI models has increased from 11\% to 96\% \citep{science-industry-numbers}. 

The relationship of computing power to success in NLP has raised concerns regarding the de-democritization and monopolization of AI \citep{de-democritization, monopoly}. Via case studies, \citet{dodge-etal-2019-show} critique benchmark reporting metrics for unfairly favouring those who have the capacity to run more experiments. The greater resources available in industry labs may give them this advantage. Our survey analyzes whether this advantage leads to more substantial benchmark improvements.

\section{Survey} \label{sec:survey}
\subsection{Affiliation} \label{sec:affil_def}

We randomly sampled 100 EMNLP 2022 papers from the main conference proceedings and classified their authors' affiliations as either academic, non-profit, or industry. Government institutes were grouped with academia as both tend to serve the general public. Non-profits were self-reported as such and typically serve a subset of the public. Lastly, any for-profit institute that primarily serves private interests and shareholders was labelled as industry.

We then compressed the group of author's affiliations into a single class. If all were from academia, the paper was labelled as such. If at least one affiliation was non-profit, the paper label was updated to non-profit. The same heuristic was used for industry with industry taking precedence over non-profit. Our rationale was that if even one author was from industry or a non-profit, the team would likely have access to the resources of that institute. We replicated our main results with affiliation class determined using majority rule, with only minor variation in results (see Appendix \ref{app:majority}).

It should be noted that even when an author's affiliation is academic, this does not preclude them from having industry funding. We limited our analysis to affiliations explicitly stated in the paper's authorship section and did not delve into acknowledgements or external CVs. Our analysis is therefore a lower bound on industry presence. The affiliation distribution of the surveyed papers is in Table \Ref{tab:paper_affils}.

\begin{table}
    \centering
    \small
    \resizebox{0.9\columnwidth}{!}{%
        \begin{tabular}{c|ccc|c}
            \toprule
            & Academia & Non-Profit & Industry & Total\\
            \midrule
            Long & 45 & 5 & 37 & 87 \\
            Short & 5 & 1 & 7 & 13 \\
            \midrule
            Total & 50 & 6 & 44 & 100 \\
            \bottomrule
        \end{tabular}}
    \caption{Affiliations of Surveyed EMNLP 2022 Papers.}
    \label{tab:paper_affils}
\end{table}

Corroborating the trend in \citep{abdalla-etal-2023-elephant}, 44\% of the surveyed EMNLP 2022 papers have industry affiliations, up from 24\% over 2019-2021.

\subsection{Type} \label{sec:paper_type}

We assigned each paper one or more type(s), most of which are self-explanatory. The type of paper can be thought of as the contribution(s) the authors are using to argue that their work is worthy of publication. Unique Setting SOTA is distinct from All-Time in that the authors are only looking at a constrained version of a dataset/benchmark. Examples include zero/few-shot, parameter-efficient tuning, and out-of-distribution robustness. Interpretability \& Analysis papers dissect a pre-existing work or method to uncover additional insights. Split by institute type, Table \Ref{tab:paper_types} presents the types of papers surveyed. 

Despite widespread criticism, performance on benchmarks remains critical to publishing in top venues with 73 SOTA claims appearing in the 100 papers. Industry also focuses on New Datasets. \citet{bowman-dahl-2021-will} detail the prohibitive cost of compiling and annotating an NLP dataset, likely making it a more viable endeavour for an industry budget. Likewise, industry produced over twice as many new Pre-Trained Models (PTMs). Academia focuses more on Interpetability \& Analysis work---an inherently computationally conservative research direction. 

Despite resource constraints, academia and industry published near identical quantities of both types of SOTA papers. One might expect the restricted environment of the Unique Setting to favour academia, both by limiting competition and possibly reducing computational requirements (e.g., parameter-efficient tuning), but this was not the case in our sample. A distinction between industry and academic SOTA advancements is, however, identifiable in terms of relative score improvement (see \cref{sec:degree}).

\begin{table}
    \centering
    \small
    \resizebox{1\columnwidth}{!}{%
        \begin{tabular}{lccc|c}
            \toprule
            Type & Academia & Non-Profit & Industry & Total \\
            \midrule
            Unique Setting SOTA & 18 & 4 & 17 & 39 \\
            All-Time SOTA & 18 & 0 & 16 & 34 \\
            New Dataset & 10 & 1 & 16 & 27 \\
            Interpretability \& Analysis & 11 & 1 & 5 & 17 \\
            New Pre-Trained Model & 2 & 0 & 5 & 7 \\
            New Metric & 1 & 0 & 5 & 6 \\
            Ethics & 2 & 0 & 1 & 3 \\
            \bottomrule
        \end{tabular}}
    \caption{Types of Surveyed EMNLP 2022 Papers. Note that the columns do not sum to the amount of papers surveyed from each institution type since papers may be assigned more than one type.}
    \label{tab:paper_types}
\end{table}

\subsection{Citations} \label{sec:citation_def}

Research is an iterative collaborative effort that continually builds upon prior work. We recorded the cited works within each surveyed paper that have the most bearing on research direction and publication validity: PTMs, Datasets, Prior Best (for both All-Time and Unique Setting SOTA), and Full Setting (distinct to Unique Setting SOTA). For example, if a paper used the GLUE benchmark \citep{wang-etal-2018-glue}, we appended it to the list of that paper's cited datasets. For each of these citations, we used the same heuristic as in \cref{sec:affil_def} to assign an affiliation label. Citations in appendices were excluded.

In addition to directly cited PTMs, if a PTM's weights were initialized from an ancillary PTM, its citation was included as well. This recursive pattern was repeated until a PTM with weights initialized from scratch was identified. We ignored citations for model architectures since the architecture itself may not be computationally prohibitive while pre-training it often is. We did so to focus the survey on works that may not be computationally feasible for all researchers. 

All datasets, whether used for pre-training, fine-tuning, and/or scoring were recorded. When there was a resultant metric from training on and/or testing on a dataset, a Prior Best (and possible Full Setting) citation (if it existed) was associated with that specific metric. Multiple metrics may have been reported on a single dataset sometimes resulting in different Prior Bests being associated with the same dataset. 

For Unique Setting SOTA papers, the Full Setting is often included in the results section to provide an upper bound for comparison and we recorded these as well (e.g. the baseline of 100\% parameters tuned in parameter-efficient tuning). Our goal was to determine which institute type usually owns this upper bound. 

For both Prior Best and Full Setting citations, we recorded the reported score associated with them. We use them to determine degree of improvement for SOTA claims (see \cref{sec:degree}).

\subsection{Miscellaneous} \label{sec:misc}

To round out our survey, we recorded two miscellaneous variables: public code release and number of recorded metrics.

Reproducibility is a common but unevenly applied criterion for credibility in NLP and scientific research as a whole. \citet{repro-ai} grouped factors that assist in reproducing AI results into three categories and from surveying 400 research papers, determined that only 20--30\% of them document such factors. One of these factors is the release of implementation code and data, a recommendation supported by \citet{dodge-etal-2019-show} and included in the EMNLP 2022 Reproducibility Criteria. If a paper provided a link to their code and/or data (typically as a GitHub repository) the public code release variable was set to \texttt{True}.

In parallel to the citation recording process, we calculated the number of reported metrics per paper to determine whether industry's increased computing capabilities would allow them to disseminate more results. This variable, along with the other miscellaneous variables are averaged over each institute type and presented in Table \Ref{tab:misc}. 

The percentage of publicly released code exhibits a downward trend across institution type. Given that reviewers were instructed to take into consideration a submission's ability to meet Reproducibility Criteria, the acceptance of 15 papers with private implementations indicates that either these papers contribute exceptional methods that outweigh the negative consequences of their privacy or that the value placed on reproducibility is not uniform across reviewers. The usefulness of scientific results reported on closed models has been called into question \cite{rogers-etal-2023-closed} and whether research on private models is suitable for a public venue such as EMNLP is worthy of debate. Note that an "Industry Track" exists at EMNLP, which would seem like a natural fit for privately-implemented industry papers, however, they were submitted to the main track instead. We reflect further on the purpose of an Industry Track in \cref{sec:discussion}. 

Academics and industry practitioners report nearly the same number of metrics per paper. This could be a result of limiting our survey to the main bodies of papers which have a fixed length and subsequently a relatively standard number of tables and figures. A future analysis including appendices may be more revealing. 

\begin{table}
    \centering
    \small
    \resizebox{0.95\columnwidth}{!}{%
        \begin{tabular}{l|ccc|c}
            \toprule
            & Academia & Non-Profit & Industry & All\\
            \midrule
            %SST Attempt & 20.0\% & 16.7\% & 36.4\% & 27.0\% \\
            Public Code & 94.0\% & 83.3\% & 75.0\% & 85.0\% \\
            Num. Metrics & 14.0 & 4.5 & 15.5 & 14.0 \\
            \bottomrule
        \end{tabular}}
    \caption{Miscellaneous survey elements averaged over all papers of each institution type.}
    \label{tab:misc}
\end{table}

\section{Analysis} \label{sec:analysis}

Some research questions were not immediately answerable from the raw survey data and are addressed in this section via additional analysis.

\subsection{Reliance on Industry Artifacts and Contributions} \label{sec:reliance}

Having recorded the affiliations of each paper's most important citations split by type (\cref{sec:citation_def}), we processed the data by binning citations by publication year. We did not split publications by the EMNLP 2022 paper's affiliation for this analysis. We wanted to quantify reliance on industry for the entire NLP community. For each year with at least one citation, we determined what percentage of the citations were industry. Some years---especially the earlier ones---had no or few citations and the sparsity resulted in jagged impulses. We applied a 1D Gaussian filter\footnote{SciPy Implementation (v1.10.1): \url{https://docs.scipy.org/doc/scipy-1.10.1/reference/generated/scipy.ndimage.gaussian_filter1d.html}} with its standard deviation, $\sigma$ set to $2.5$ to smooth the data and isolate a long-term trend. 

From \citet{abdalla-etal-2023-elephant} we have access to the actual ACL Anthology industry affiliation rate over time and we plot it alongside the citation data. We label this line "expected" since one would expect the citations from each year to follow the same trend. Note that the ACL Anthology is not the publisher of every citation in EMNLP 2022. Image PTMs for example are typically published at other venues. However, the trend of increased industry presence at ACL generalizes to other major AI conferences \citep{de-democritization} and given that most NLP publications cite within the NLP community \cite{wahle2023we} we believe it to be an adequate proxy of expectation.

The plot of both our citation data and expected industry presence is in Figure \ref{fig:years}. Two observations are immediately clear: The citation of industry papers is well above expected for every year and the  proportion is increasing. In fact, for Datasets, PTMs, Prior Bests, and Full Setting citations as of 2022, the majority are industry. This trend indicates a reliance on industry artifacts and contributions. While only 13.3\% of published ACL Anthology papers were from industry in 2022, over three times that proportion were used as important citations. 

In terms of artifacts, having the most used Datasets and PTMs coming from industry may limit the broader NLP community's research directions to align with the publishing industry. Table \ref{tab:no_ptm} shows that while there is no explicit requirement that EMNLP papers use a PTM, the actuality that nearly all papers do implies otherwise. Since nearly all PTMs are produced by industry, it being a \textit{de facto} standard to use them means that nearly all NLP research includes an industry component. %Since nearly all PTMs are produced by industry, further reliance on industry is perpetuated and a picture of corporate capture becomes clearer. 

Industry's majority claim on Prior Bests suggests that although they published less in the past, their publications are more likely to achieve and retain SOTA. The situation for Full Setting citations also supports a trend of industry reliance where the majority of Unique Setting works are at their onset looking up to industry as an impractical upper bound.

\begin{table}
    \centering
    \small
    \resizebox{0.9\columnwidth}{!}{%
        \begin{tabular}{l|ccc|c}
            \toprule
            PTM & Academia & Non-Profit & Industry & Total\\
            \midrule
            Yes & 43 & 6 & 38 & 87 \\
            No & 7 & 0 & 6 & 13 \\
            \midrule
            Percent & 86 & 100 & 86 & 87 \\
            \bottomrule
        \end{tabular}}
    \caption{Quantities of papers that used at least one PTM.}
    \label{tab:no_ptm}
\end{table}

For an analysis on how surveyed papers were predisposed to citing papers from within their own institute type, see Appendix \ref{app:predis}.

% \footnotetext{SciPy Implementation (v1.10.1): \url{https://docs.scipy.org/doc/scipy-1.10.1/reference/generated/scipy.stats.ttest_ind.html}}

\subsection{Degree of SOTA Improvement} \label{sec:degree}

\begin{figure}
    \centering
    \resizebox{\linewidth}{!}{
        \includegraphics{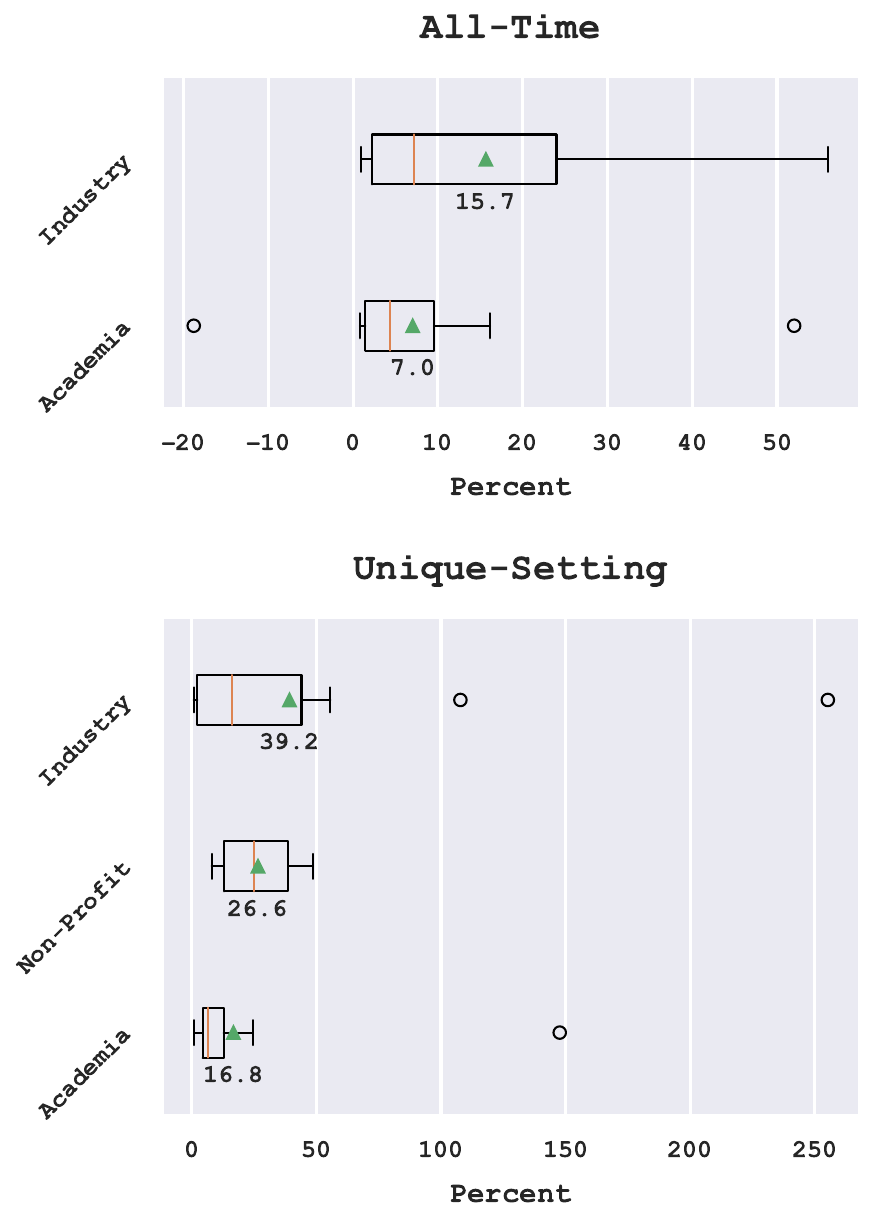}
    }
    \caption[Caption for LOF]{Per paper average relative score increase distributions grouped by institution type for  All-Time (top) and Unique (bottom) Setting SOTA papers. No non- profit papers surveyed claimed All-Time SOTA and the corresponding row is therefore excluded. Green triangles and orange vertical bars denote the mean and median respectively and the numerical value for the mean is labelled beneath it.}
    \label{fig:rel_comb_scores}
\end{figure}

There was no clear distinction between quantities of SOTA papers published by industry and academia (\cref{sec:paper_type}). However, by analyzing the distribution of score increases over prior bests, a difference between institution types became evident. To do so, we averaged the relative improvements of individual metrics first over the dataset they were scored on, and then again over all datasets evaluated per paper. Equation \ref{eq:rel} formalizes our approach:

\begin{equation} \label{eq:rel}
    \%\Delta_p = \frac{1}{D}\sum_{d=1}^{D}\frac{1}{M_d}\sum_{m=1}^{M_d}\frac{new_{dm} - old_{dm}}{old_{dm}}
\end{equation}
where $D$ is the number of datasets a paper evaluated on, $M_d$ is the number of metrics per dataset, $d$, $new_{dm}$ is the paper's new score reported on $d$ and metric, $m$, and $old_{dm}$ is the Prior Best score on the same metric and dataset. For metrics where lower scores are preferable $new_{dm}$ and $old_{dm}$ were swapped. Metrics with an old score of zero were automatically set to a relative increase of $100\%$ regardless of new score. The papers' average relative score increase were then grouped by institute. Outliers outside two times the standard deviation of each group were removed (see Appendix \ref{app:outliers}) and the remaining points are plotted in Figure \ref{fig:rel_comb_scores}. 

\begin{table*}
    \centering
    \small
    \resizebox{\textwidth}{!}{%
        \begin{tabular}{ccccccc}
            \toprule
            Rank & Score Inc. & Institute Type & SOTA Type & Paper Topic &  Avg. Year of PB & Avg. Num. PBs \\ 
            \midrule
            
            1 & 255\% & Industry & Unique & Multilingual PTM & 2020 & 1 \\
            2 & 148\% & Academia & Unique & Retrieval-Based Dialogue Multi-View Response Selection & 2019 & 9 \\
            3 & 108\% & Industry & Unique & Continual Learning & 2019 & 1 \\
            4 & 56.0\% & Industry & All-Time & Written and Spoken Long Document Summarization & N/A\footnotemark & 6 \\
            5 & 55.6\% & Industry & Unique & Aspect-Based Sentiment Analysis & 2021 & 2 \\
            6 & 52.0\% & Academia & All-Time & Knowledge Graph Completion & 2018 & 2 \\
            7 & 48.6\% & Non-Profit & Unique & Abstractive Text Summarization & 2022 & 5 \\
            8 & 44.9\% & Industry & Unique & Controllable Text Summarization & 2019 & 1 \\
            9 & 43.3\% & Industry & Unique & Generative Language Decoding & 2020 & 4 \\
            10 & 42.2\% & Industry & All-Time & Speaker Overlap-aware Neural Diarization & 2022 & 3 \\
            \midrule

            58 & 1.16\% & Industry & Both & Bilingual Lexicon Induction & 2019 & 4 \\
            59 & 1.12\% & Academia & Unique & Parameter Efficient Training & 2021 & 2 \\
            60 & 1.06\% & Industry & Unique & Transformer Model Compression & 2020 & 1 \\
            61 & 1.04\% & Academia & All-Time & Aspect Sentiment Triplet Extraction & 2021 & 6 \\
            62 & 1.00\% & Industry & Unique & Factual Consistency in Summarization & 2021 & 3 \\
            63 & 0.99\% & Industry & All-Time & Multi-Domain Machine Translation & 2020 & 5 \\
            64 & 0.89\% & Academia & All-Time & Text Style Transfer & 2020 & 4 \\
            65 & 0.84\% & Industry & Unique & Machine Translation & 2022 & 1 \\
            66 & 0.81\% & Academia & All-Time & Aspect Sentiment Triplet Extraction & 2022 & 7 \\
            67 & -18.8\% & Academia & All-Time & Controllable Text Generation & 2021 & 6 \\
            
            \bottomrule
        \end{tabular}}
    \caption{Top and bottom 10 Per Paper Average Relative Score Increase (\cref{sec:degree}). PB stands for prior best. Average number of prior bests is how many were compared to within the paper per metric, averaged over all metrics.}
    \label{tab:qual}
\end{table*}

The means of industry SOTA improvements are over two times times higher than academia's for both All-Time and Unique Setting. Academia's negative data-point indicates that some academics are claiming SOTA despite failing to achieve a net score increase (more on this in \cref{sec:qual}). Requirements to claim SOTA are not strictly defined and some metrics may be weighted higher than others in a paper, but this finding along with the overall lower means, suggests that academia's SOTA claims are not as strong as industry's. Although there were equal numbers of SOTA claims between institution types, academia's weaker SOTA claims, as well as the disproportion of prior bests being from industry (\cref{sec:reliance}), could result in a gradual exclusion of academia from future SOTA claims. 

Unique Setting claims are more substantial than All-Time. By narrowing the problem to a Unique Setting, competition is automatically removed and it logically follows that higher score increases would be achievable. This pattern is noticeable for both academia and industry. The smaller advancements for All-Time claims supports the theory that benchmarks are saturated \citep{bowman-dahl-2021-will}. There is less room for researchers to maneuver at the top of the leaderboards, decreasing the strength of new SOTA claims. 

Relative score is biased toward increases where the old score is relatively low; a phenomenon more common to the Unique Setting. To contrast relative increase's inherent bias we also plotted absolute score increase in Appendix \ref{app:abs}. In this case, academia captures SOTA equally as effectively as industry. Without the benefit of improving over small prior scores, industry's apparent advantage disappears. This pattern could suggest that industry prefers or is more capable of addressing novel tasks (i.e. ones without much prior success, and hence low old scores) while academia sticks to safer, well-established benchmarks. 

\subsection{Qualitative Analysis} \label{sec:qual}

\footnotetext{Prior bests in this paper were not always cited and therefore an average year could not be determined.}

Individual SOTA papers with the greatest and lowest average relative score increases (see \cref{sec:degree} for formalization) are listed in Table \ref{tab:qual}. Only two of the top ten score increases are from academia despite the majority of surveyed papers being academic. Papers ranked 1, 3, and 9 all introduce new PTMs, none of which are from academia, possibly indicating that training a PTM is both more likely to produce impressive results and that industry is more likely to succeed at it. 

The worst relative score improvement is -18.8\% and comes from academia. This paper does not fall outside of the threshold for outliers (see Appendix \ref{app:outliers}) and therefore cannot be ignored. On nearly every metric reported, the paper's method performed worse than a prior best. Nevertheless, the authors still explicitly claim SOTA in the abstract. The paper proposes a novel method worthy of publication but an erroneous SOTA claim should not have been used to argue its validity. This could be an example of authors succumbing to the discourse within NLP of "reject if not SOTA" \citep{reject-not-sota}. Only one of the four prior bests they are closest to are academic, indicating that this could be an area saturated with industry competition. 

We recorded the final two columns of Table \ref{tab:qual} to discern whether there is a distinction in prior best recency and quantities between the highest and lowest SOTA-achieving papers. The top 10 SOTA-increasing papers’ prior bests have an average publication date of January 2020 versus August 2020 for the bottom 10. The top 10 compare their new results to an average of 3.4 prior bests compared to 3.9 for the bottom 10. These statistics follow the logic that it would be easier to achieve higher SOTA gains when comparing to older and lower prior bests.

\section{Discussion} \label{sec:discussion}

Industry presence in NLP artifacts and contributions is common and well-above the expected margin. From the perspective of policy research and those who favour research productivity, this relationship may be seen as a purely positive collaboration \citep{gulbranson-ind-funding-and-performance, reconciling-anti-pro-industry}. Industry has committed substantial financial resources to NLP and the entire research community benefits from more funding and greater professional demand \citep{index-report}. It could be argued that the progress achieved in recent years would not have been possible without a high degree of industry involvement.

In contrast, those who consider objectivity as an important epistemic virtue of scientific research may be concerned with this trend. \citet{abdalla-grey-hoodie} compare the state of AI Ethics to the tobacco industry at its peak, a situation which in retrospect is usually cast in a negative light. In a few decades time, will we look back at NLP and AI research with similar collective disdain? Will our present efforts be overshadowed by a narrative of objectivity tainted by corporate capture? At the very least, whether for good or bad, publishing in top NLP venues such as EMNLP without using industry artifacts and contributions is becoming increasingly infeasible.

Regardless of viewpoint, all stakeholders relevant to this discussion benefit from increased transparency. Throughout our survey we identified areas that our research community could address to assist in quantifying industry presence and reducing barriers to publication. We suggest the following: 

\begin{itemize}
    \item Authors should be required to list all sources of funding for their work. Without specifics, we had to rely on author affiliation to quantify industry presence. This is an imprecise method given that academia, non-profits, and for-profits all receive some proportion of public funding. Requiring an acknowledgement section listing \textit{all} sources of funding would alleviate this issue without excessive overhead.
    
    \item In line with the above suggestion, tagging papers according to public, private, and/or a split of funding beside its link in the ACL Anthology would foster transparency. The tag could look similar to dataset, software, and best paper tags as in EACL 2023\footnote{\url{https://aclanthology.org/events/eacl-2023/}} and link to descriptions of each category of funding. 

    \item There already exists an "Industry Track" at *ACL conferences that calls for papers related to "non-trival real-world systems". However, authorship is not restricted to for-profit industry nor is industry excluded from the main track. This version of an Industry Track is misleading in that its title implies a separation by institute type, when in practice that sepration is non-existent. Future works should look into what inclusions and exclusions could be added to the Industry Track for it to better serve its expected purpose. For example, should private implementations be only acceptable in this track?
    
    \item Reserving a fixed proportion of publication slots for various hierarchies of computing capabilities could serve as an initial idea to equitably address the compute divide \citep{de-democritization}. Coincidentally, doing so would help address the transparency concerns of evaluating on single scores \citep{dodge-etal-2019-show}. How to disclose computing capabilities is not immediately obvious and we leave that to future work.

    \item To support the validity of non-SOTA papers, *ACL conferences could reserve publication slots for them. The ACL 2023 Review Policy \citep{acl-2023-frontmatter} discouraged review shortcuts such as "reject if not SOTA" but this discourse is pervasive and explicit support of non-SOTA papers may be necessary to overcome it. 
\end{itemize}

These suggestions each share the common goal of increasing transparency regarding funding sources and addressing the gap in computing resources between institutions. The compute divide is a well-understood phenomenon and yet it continues to be ignored when submitting and reviewing publications. Researchers with less funding are excluded from a subset of problems due to insufficient computing resources. They can only gain access by collaborating with industry, which may not align with their values or abilities\footnote{For example, universities without globally-recognized prestige may have a harder time attracting industry sponsorship.}. Our main concern is ensuring that a space always exists for all NLP practitioners to contribute their work. Preserving institutional diversity in NLP will strengthen our collective research by ensuring that perspectives from stakeholders without conflicts of interest with industry are still heard.

\section{Conclusion}

Industry has increased investment in AI research and specifically NLP in recent years. Our survey of 100 EMNLP 2022 papers found that the citation of industry artifacts and contributions is far greater than what would be expected based on yearly industry publication rates. This relationship indicates a reliance on industry. We conclude with a brief discussion highlighting possible positive and negative impacts of industry reliance and provide suggestions for navigating transparency and computing equity issues in NLP.

\section*{Limitations}

\subsection*{Sample Size}
The most obvious limitation of our analysis is sample size. We studied a starting sample of 100 out of a possible 829 papers. Reading each paper and manually extracting the relevant citations (and reading those papers as well) was time consuming. We considered performing the data collection automatically with NLP solutions, but doing so would have likely required a pre-existing labelled dataset. Using the data we released with this paper for training an NLP model to do this type of analysis is an interesting line of future work. Regardless, our main results that identify a reliance on industry artifacts and citations (Figure \ref{fig:years}) make use of several citations within each surveyed paper such that the sample size is far greater than 100. For exact numbers see Appendix \ref{app:predis} where the samples sizes for each citation type are reported in Figure \ref{fig:bars} as ``$n$''.

\subsection*{Defining Institution Types}
It is difficult to bin authors into clearly defined institute types from affiliation alone. Academics receive funding from industry scholarships and grants and conversely industry can be the beneficiary of public funding (in the forms of grants, tax rebates, credits, etc.). Additionally, some academic institutes are classified as private entities. The blurred line between public and private research makes disentangling them nearly impossible. Our analysis is therefore only an estimate based on information clearly presented in the authorship sections of papers. The implementation of our suggestion in \cref{sec:discussion} regarding reporting funding sources explicitly in the Acknowledgements section of papers or something similar will improve precision in future studies.

\subsection*{Proving Capture}
It is difficult for researchers to prove that the increased reliance demonstrated through our analysis is proof of corporate capture instead of mutually beneficial collaborations. That is, to formally prove `Science Capture', we would need to demonstrate that this increased reliance has changed the research questions that would have been asked (without said reliance), or that the way research results and topics are discussed would have changed without the observed interactions. Without personal admissions from industry leaders, capture can be measured by diachronic analysis of NLP over time comparing research trajectories in eras with large and minimal industry presence (or different NLP subfields with various levels of industry presence). We hope future research can tackle this research goal.

\section*{Ethics Statement}
The information we collect and release in our survey is publicly available in the EMNLP 2022 Conference Proceedings\footnote{\url{https://aclanthology.org/events/emnlp-2022/}} under a Creative Commons Attribution 4.0 International License and therefore did not qualify for internal ethics review. Furthermore, we do not identify specific papers by name in our work. However, it is likely possible to sort through the EMNLP 2022 proceedings to identify papers in Table \ref{tab:qual}. We discourage this practice. Our intention was to analyze patterns regarding broad institution types and not to call into question the contributions of individual scientists.

\section*{Acknowledgements}
For this research, WA was funded by a Natural Sciences and Engineering Research Council of Canada (NSERC) Canadian Graduate Scholarship - Master's (CGS-M), KR was funded by an NSERC Discovery Grant, and CS was funded by the Office of the Privacy Commissioner of Canada; the views expressed herein are those of the authors and do not necessarily reflect those of the funders.

\bibliography{anthology,custom}
\bibliographystyle{acl_natbib}

\appendix

\section{Majority Affiliation Labelling} \label{app:majority}

In this section we re-report results using the majority institute type of authors to classify a paper as Academic, Non-Profit, or Industry. These results echo the findings of the original labelling heuristic of \cref{sec:affil_def} albeit less pronounced.

\begin{table}[hb]
    \centering
    \small
    \resizebox{0.9\columnwidth}{!}{%
        \begin{tabular}{c|ccc|c}
            \toprule
            & Academia & Non-Profit & Industry & Total\\
            \midrule
            Long & 66 & 4 & 17 & 87 \\
            Short & 6 & 1 & 6 & 13 \\
            \midrule
            Total & 72 & 5 & 23 & 100 \\
            \bottomrule
        \end{tabular}}
    \caption{Affiliations of Surveyed EMNLP 2022 Papers using majority labelling.}
\end{table}

\begin{table}[h]
    \centering
    \small
    \resizebox{1\columnwidth}{!}{%
        \begin{tabular}{lccc|c}
            \toprule
            Type & Academia & Non-Profit & Industry & Total \\
            \midrule
            Unique Setting SOTA & 26 & 3 & 10 & 39 \\
            All-Time SOTA & 29 & 0 & 5 & 34 \\
            New Dataset & 16 & 1 & 10 & 27 \\
            Interpretability \& Analysis & 14 & 1 & 2 & 17 \\
            New Pre-Trained Model & 4 & 0 & 3 & 7 \\
            New Metric & 3 & 0 & 3 & 6 \\
            Ethics & 2 & 0 & 1 & 3 \\
            \bottomrule
        \end{tabular}}
    \caption{Types of Surveyed EMNLP 2022 Papers, counted using majority labelling.}
\end{table}

\begin{table}[h]
    \centering
    \small
    \resizebox{1\columnwidth}{!}{%
        \begin{tabular}{l|ccc|c}
            \toprule
            & Academia & Non-Profit & Industry & All\\
            \midrule
            %SST Attempt & 20.0\% & 16.7\% & 36.4\% & 27.0\% \\
            Public Code & 86.1\% & 100\% & 78.2\% & 85.0\% \\
            Num. Metrics & 13.6 & 4.0 & 17.4 & 14.0 \\
            \bottomrule
        \end{tabular}}
    \caption{Miscellaneous survey elements averaged over all papers of each institution type labelled using the majority heuristic.}
\end{table}

\begin{figure}
    \centering
    \resizebox{\linewidth}{!}{
        \includegraphics{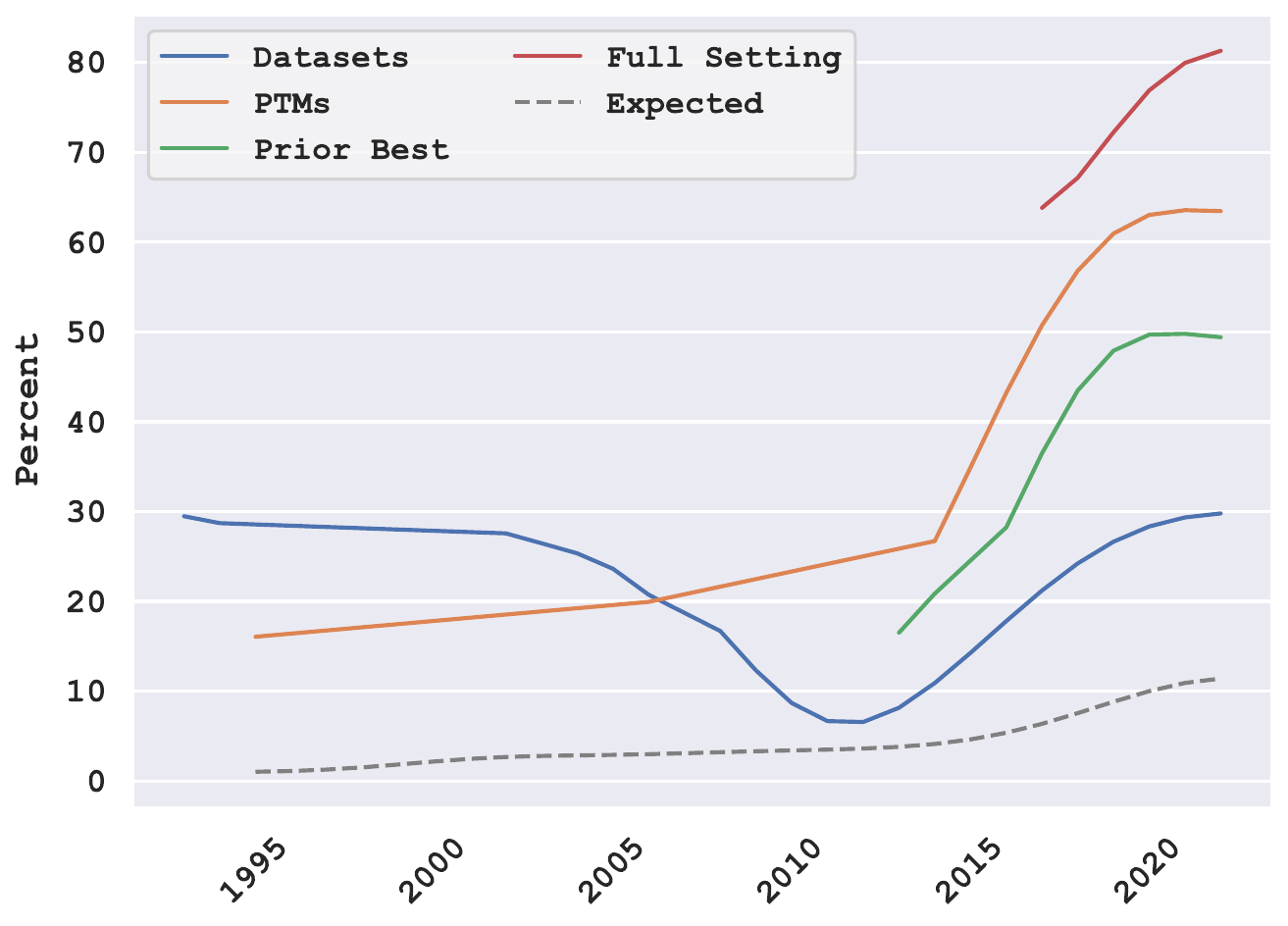}
    }
    \caption{Percentage of surveyed EMNLP 2022 paper citations with industry affiliation---labelled according to majority---smoothed with a 1D Gaussian Filter ($\sigma = 2.5$). The "Expected" line uses the original heuristic for quantifying industry as specified in \cref{sec:affil_def} since that is how \citet{abdalla-etal-2023-elephant} reported.} 
\end{figure}

\begin{table}
    \centering
    \small
    \resizebox{1\columnwidth}{!}{%
        \begin{tabular}{l|ccc|c}
            \toprule
            PTM & Academia & Non-Profit & Industry & Total\\
            \midrule
            Yes & 63 & 5 & 19 & 87 \\
            No & 8 & 0 & 5 & 13 \\
            \midrule
            Percent & 88.7 & 100 & 79.2 & 87 \\
            \bottomrule
        \end{tabular}}
    \caption{Quantities of majority-labelled papers that used at least one PTM.}
\end{table}

\begin{figure}
    \centering
    \resizebox{\linewidth}{!}{
        \includegraphics{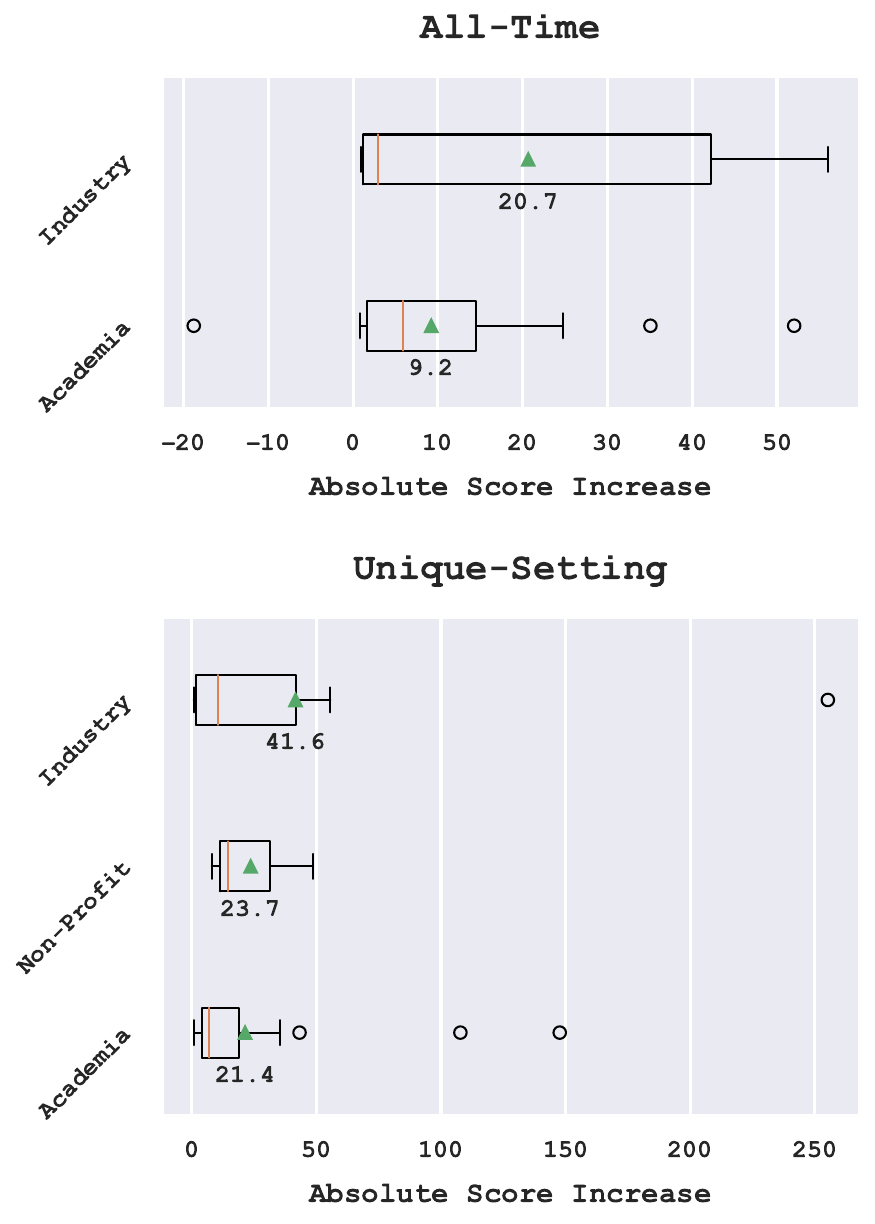}
    }
    \caption{Per paper average relative score increase distributions grouped by institution type---labelled via majority heuristic---for All-Time (top) and Unique (bottom) Setting SOTA papers. Green triangles and orange vertical bars denote the mean and median respectively and the numerical value for the mean is labelled beneath it.}
\end{figure}

\begin{table*}
    \centering
    \small
    \resizebox{\textwidth}{!}{%
        \begin{tabular}{ccccccc}
            \toprule
            Rank & Score Inc. & Institute Type & SOTA Type & Paper Topic &  Avg. Year of PB & Avg. Num. PBs \\ 
            \midrule
            
            1 & 255\% & Industry & Unique & Multilingual PTM & 2020 & 1 \\
            2 & 148\% & Academia & Unique & Retrieval-Based Dialogue Multi-View Response Selection & 2019 & 9 \\
            3 & 108\% & Academia & Unique & Continual Learning & 2019 & 1 \\
            4 & 56.0\% & Industry & All-Time & Written and Spoken Long Document Summarization & N/A & 6 \\
            5 & 55.6\% & Industry & Unique & Aspect-Based Sentiment Analysis & 2021 & 2 \\
            6 & 52.0\% & Academia & All-Time & Knowledge Graph Completion & 2018 & 2 \\
            7 & 48.6\% & Non-Profit & Unique & Abstractive Text Summarization & 2022 & 5 \\
            8 & 44.9\% & Industry & Unique & Controllable Text Summarization & 2019 & 1 \\
            9 & 43.3\% & Academia & Unique & Generative Language Decoding & 2020 & 4 \\
            10 & 42.2\% & Industry & All-Time & Speaker Overlap-aware Neural Diarization & 2022 & 3 \\
            \midrule

            58 & 1.16\% & Industry & Both & Bilingual Lexicon Induction & 2019 & 4 \\
            59 & 1.12\% & Academia & Unique & Parameter Efficient Training & 2021 & 2 \\
            60 & 1.06\% & Academia & Unique & Transformer Model Compression & 2020 & 1 \\
            61 & 1.04\% & Academia & All-Time & Aspect Sentiment Triplet Extraction & 2021 & 6 \\
            62 & 1.00\% & Industry & Unique & Factual Consistency in Summarization & 2021 & 3 \\
            63 & 0.99\% & Industry & All-Time & Multi-Domain Machine Translation & 2020 & 5 \\
            64 & 0.89\% & Academia & All-Time & Text Style Transfer & 2020 & 4 \\
            65 & 0.84\% & Industry & Unique & Machine Translation & 2022 & 1 \\
            66 & 0.81\% & Academia & All-Time & Aspect Sentiment Triplet Extraction & 2022 & 7 \\
            67 & -18.8\% & Academia & All-Time & Controllable Text Generation & 2021 & 6 \\
            
            \bottomrule
        \end{tabular}}
    \caption{Top and bottom 10 Per Paper Average Relative Score Increase (\cref{sec:degree}). In this version of the table, institute type is determined with majority labelling. PB stands for prior best. Average number of prior bests is how many were compared within the paper per metric, averaged over all metrics.}
\end{table*}

\clearpage

\section{Citation Predisposition} \label{app:predis}

\begin{figure*}
    \centering
    \resizebox{\linewidth}{!}{
        \includegraphics{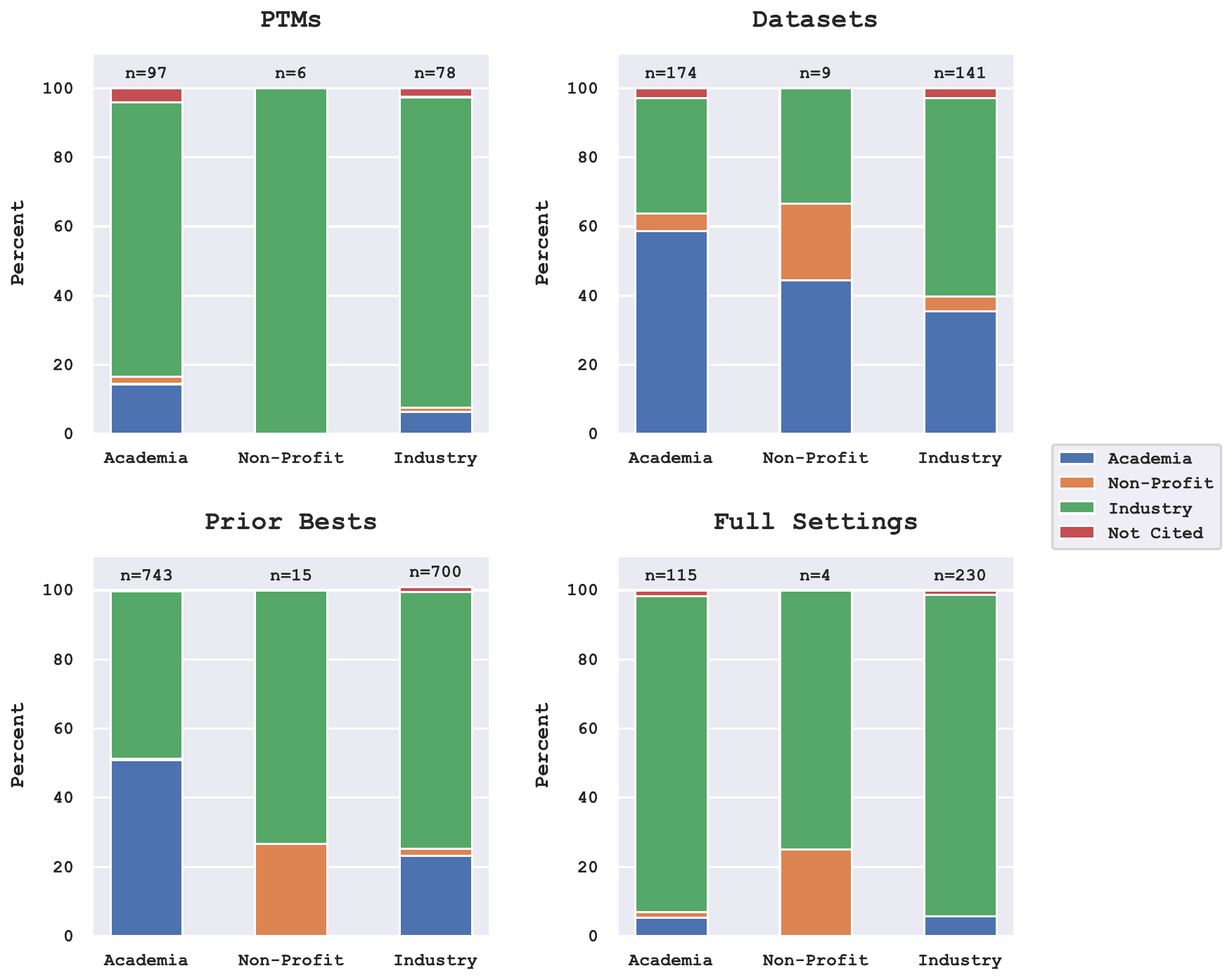}
    }
    \caption{Proportion of cited institute types split by EMNLP 2022 author affiliation. Citations are grouped by PTM (upper left), Datasets (upper right), Prior Bests (bottom left), and Full Settings (bottom right). $n$ is the total number of citations per bar. The sum of $n$ per subplot exceeds the 100 papers examined since each paper often cites more than one PTM or Dataset and reports more than one score for Prior Bests and Full Settings.}
    \label{fig:bars}
\end{figure*}

We have established that there is an overall reliance on industry across publications, but still wanted to determine whether the reliance is equal across academia and industry. Figure \ref{fig:bars} plots the same data as in Figure \ref{fig:years} but split by publishing institute type instead of citation year.

With the exception of academic citations of Full Settings, for PTMs, Datasets, Prior Bests, and Full Settings, authors are predisposed to citing their own institute type (i.e. for each of those subplots, the largest bar for each cited institute type falls within that same publishing institute type). This result is unsurprising. Researchers are more likely to address problems relevant to their institute and use a similar subset of artifacts to model and evaluate. 

Regardless of predisposition, all groups are still overwhelmingly reliant on industry PTMs. Table \ref{tab:no_ptm} shows that the alternative of not using any PTM at all is largely infeasible with only 13 papers publishing without them. Even in this scenario, there may still be a dependency on computing power if the authors are training a large model from scratch.

\section{Outliers} \label{app:outliers}

Three papers were excluded from the relative score analysis in \cref{sec:degree} with increases of $2067\%$, $2150\%$, $3246\%$. Two were from industry and the other was from academia. The first score resulted from improving a privacy preserving algorithm from near constant leakage to near perfection. The second result was due to a comparison of a baseline that performed close to zero for few-shot and zero-setting, a scenario that naturally inflates relative score increase. The final score was boosted from comparing unsupervised setting results to a method that was not designed for that setting and correspondingly performed poorly. These three were not outliers in the absolute score analysis, emphasizing the balancing effect of using both analyses. 

\section{Absolute Score Improvement} \label{app:abs}

\begin{figure}
    \centering
    \resizebox{\linewidth}{!}{
        \includegraphics{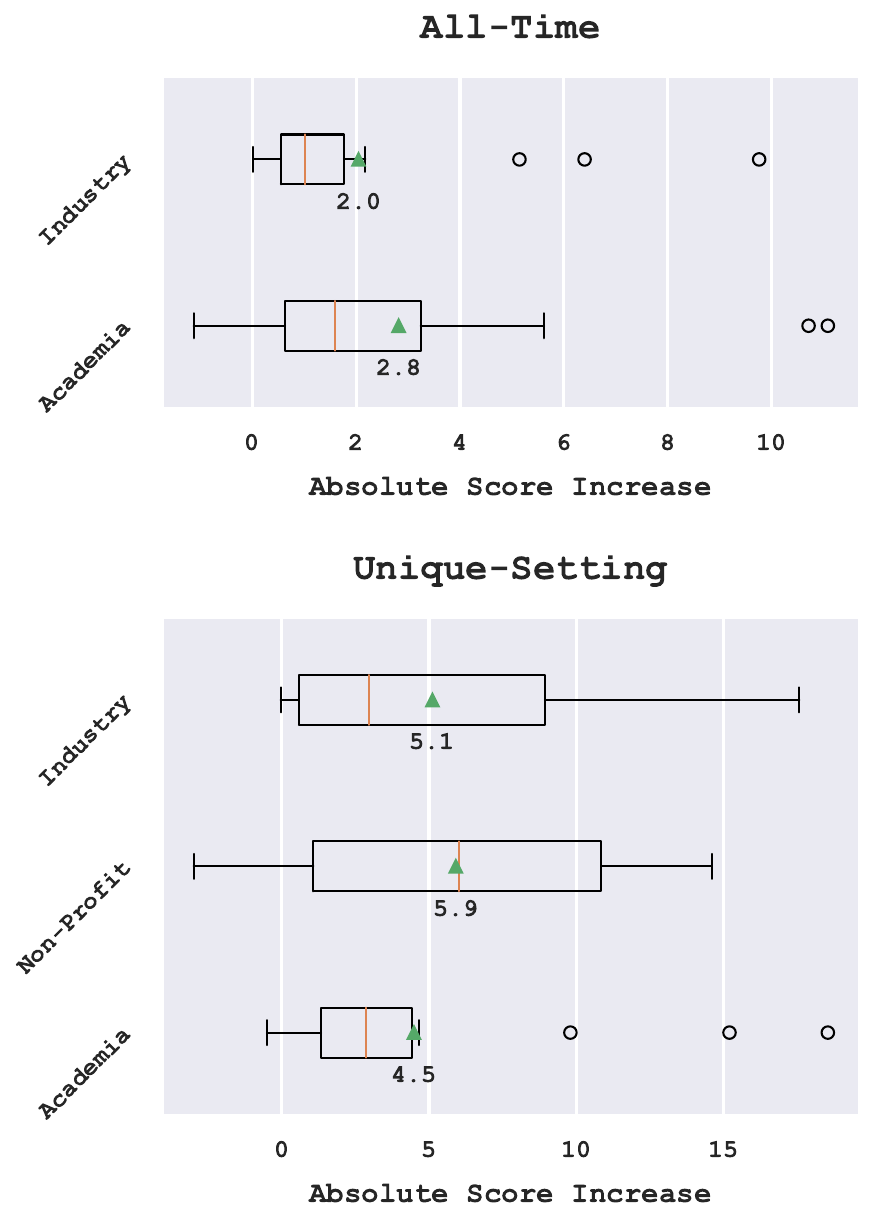}
    }
    \caption{Per paper average aboslute score increase distributions grouped by institution type for  All-Time (top) and Unique (bottom) Setting SOTA papers.}
    
    \label{fig:abs_comb_score}
\end{figure}

To contrast the bias relative score increase assigns to improvements over near-zero scores, we also plotted absolute score increase split by institution type which can be seen in Figure \ref{fig:abs_comb_score}. In this scenario, industry and academia improvements are near identical. The distinction between the two is only noticeable in terms of relative improvement.

\end{document}